\documentclass[runningheads]{llncs}
\usepackage[T1]{fontenc}
\usepackage{graphicx}
\usepackage{float}

\usepackage{amsmath,amssymb}
\usepackage{booktabs}
\usepackage{longtable}
\usepackage{hyperref}
\usepackage{url}

\usepackage{mathtools}
\mathtoolsset{showonlyrefs}

\graphicspath{{figures/}}
\title{LIG: Layer-wise Integrated Gradients for Within-Layer Flow Analysis in Transformers}
\titlerunning{Within-Layer Flow Analysis with LIG}
\author{Eight Suzuki\inst{1,3} \and
Hideitsu Hino\inst{1,2} \and
Noboru Murata\inst{1}}
\authorrunning{E. Suzuki et al.}
\institute{Waseda University\\
\email{suzuki8@akane.waseda.jp, noboru.murata@eb.waseda.ac.jp}
\and
The Institute of Statistical Mathematics\\
\email{hino@ism.ac.jp}
\and
Fujitsu Limited
}
\begin{document}
\maketitle
\begin{abstract}
Transformers achieve strong performance, but their internal computations remain opaque. We view each Transformer layer as a dynamic graph whose nodes are token representations and per-head attention outputs, with Multi-Head Attention (ATT) and MLP as module boundaries. On this graph we use LIG (Layer-wise Integrated Gradients), which applies set-to-set Integrated Gradients (IG) at nonlinear module boundaries. Set-to-set IG applies IG to a map from a set of input token representations to a set of output representations, evaluating token-to-token contributions, which is not standard in prior IG applications.
This extends IG from the usual scalar-objective setting to set-to-set maps via an $L_2$ scalarization, and composes within-layer contributions in the spirit of Layer-wise Relevance Propagation (LRP), with IG completeness playing the role of LRP-style conservation at each boundary.
We use LIG to analyze (i) the agreement between module-wise composition and layer-whole attribution under an $L_2$ criterion, and (ii) within-layer information flow by tracing separated ATT and MLP contributions. On BERT-base and PTB, configurations that best preserved within-layer consistency used the target token's embedding as the ATT baseline and either the ATT output at $a=0$ or Zero as the MLP baseline. We therefore present LIG as a diagnostic XAI tool at module-boundary granularity, without model-specific retraining or per-operation interpreter design. Code is available at \url{https://github.com/eightsuzuki/layer-wise-integrated-gradients}.
\end{abstract}

\section{Introduction}
\label{sec:introduction}

%\subsection{Background}
%\label{sec:intro-background}

The Transformer~\cite{vaswani2017} is a landmark architecture that flexibly integrates context via attention and dramatically improved NLP performance under large-scale parallel training.
Transformer information processing can be viewed as dynamically restructuring a set of input token embedding vectors into the next-stage feature vector set through the attention mechanism (in this paper, we abbreviate the within-layer block often called Multi-Head Attention (MHA) as ATT) and a multilayer perceptron (MLP).
As layers stack, weighting and integration of information proceed and dispersed representations are organized along context.
This capability spread encoder models such as BERT~\cite{devlin2019} and decoder large language models (LLMs) typified by GPT~\cite{brown2020}, while massive parameters and computation worsened opacity of prediction rationales (the black-box problem).

%Below, we write attention mechanism when describing the mechanism and ATT at module boundaries and in experimental conditions.
%We view each Transformer layer as a dynamic graph whose nodes are token representation vectors and per-head attention outputs and whose edges are module boundaries at ATT and MLP.
%The ability of edge thickness (contribution magnitude) to change with each input sentence supports the high performance that Transformers exhibit on large and diverse collections of text (Fig.~\ref{fig:intro-node-viz}).

\begin{figure}[H]
  \centering
  \includegraphics[width=0.80\linewidth]{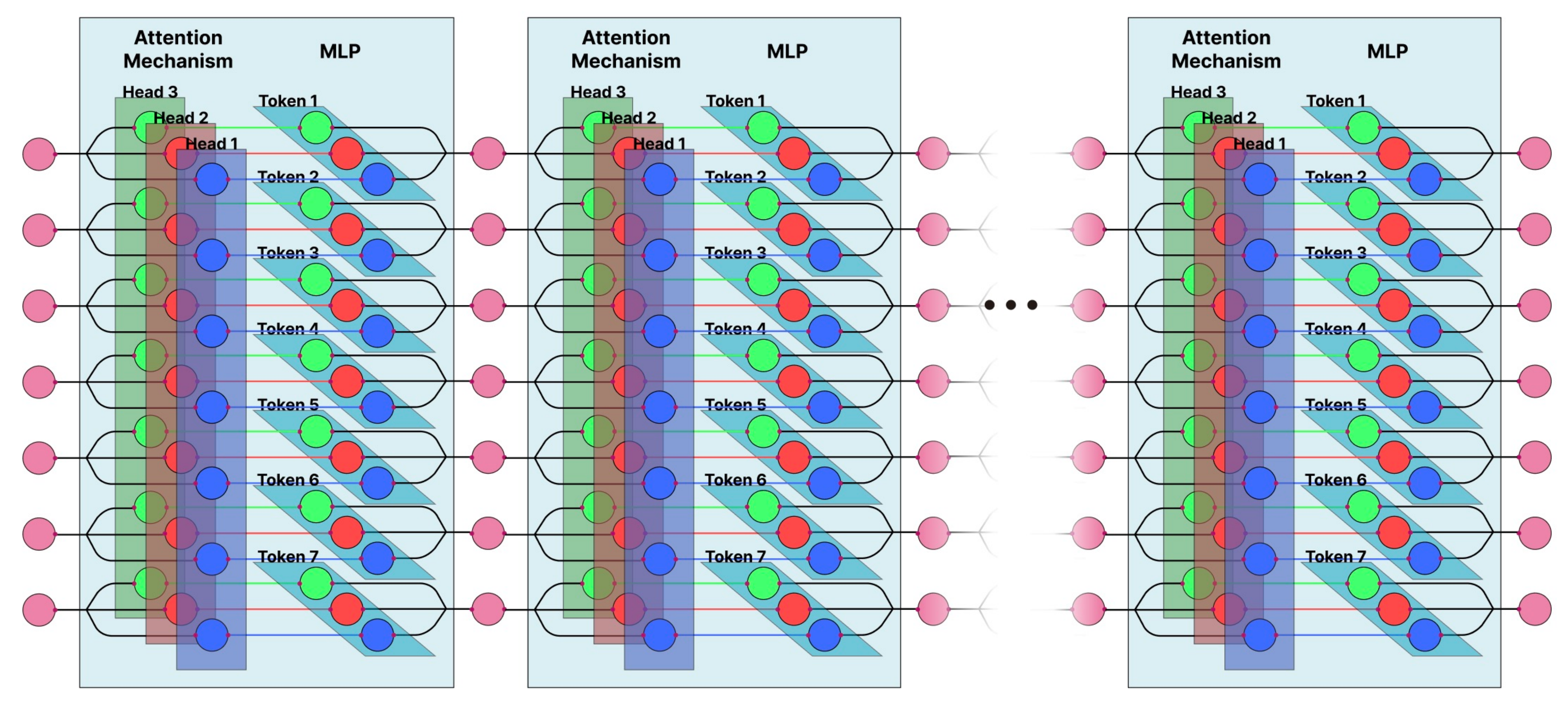}
  \caption{Basic Transformer structure and the dynamic-graph view.
  A token sequence enters from the left and passes through multiple Transformer layers (each with ATT and MLP) to become a context-aware set of vectors.
  Within each layer, token representation vectors and per-head attention outputs are nodes; information flows at ATT and MLP module boundaries.}
  \label{fig:intro-node-viz}
\end{figure}
\noindent
Most of Explainable AI (XAI) methods~\cite{arrieta2020} focus on input-to-output contributions and do not directly describe the information flow that separates ATT and MLP within the same layer.
Factor attribution methods such as SHAP are typical examples.
Methods that track internal representations exist but incur large computational and design cost for retraining interpretable models or designing propagation rules per operation (Sec.~\ref{sec:related_work}).
The contributions of this paper are twofold.
First, we apply Integrated Gradients (IG) from a set of input token representations to a set of output representations (set-to-set IG) at nonlinear module boundaries (ATT and MLP).
IG completeness at each boundary gives the same kind of conservation relation that motivates layer-wise propagation in Layer-wise Relevance Propagation (LRP), so within-layer composition can be chained without breaking that propagation view.
We further identify that inheriting the ATT output at integration endpoint $a=0$ as the MLP baseline ($\mathrm{ATTITBa}=0$), or using Zero, ranks among the best normalized $L_2$ configurations. %Under this normalized diagnostic, switching between these two MLP baselines changes $L_2$ by at most $\approx 0.001$, whereas ATT baseline design matters more.
Under this normalized diagnostic, the MLP baseline choice has little effect, whereas the ATT baseline is the dominant factor.
Second, as an application of this framework, LIG (Layer-wise Integrated Gradients) separates and composes within-layer flow between ATT and MLP, enabling macroscopic layer-wise visualization of token-to-token contributions (Fig.~\ref{fig:z2z_token_contrib_example}) and qualitative reading of how processing roles shift from local relations to sentence-wide integration across depth.

\section{Related Work}
\label{sec:related_work}

\subsection{Explainability and Post-hoc Explanation}
\label{sec:related_xai}

Explainable AI (XAI)~\cite{arrieta2020} aims to present model internal behavior and prediction rationales in forms understandable to humans.
%Arrieta et al.\ broadly distinguish 
Intrinsically interpretable transparent models are broadly distinguished from post-hoc explanations attached to trained models in~\cite{arrieta2020} .
This paper does not target the former; we treat only post-hoc explanation.

We then organize model-agnostic post-hoc explanations that construct explanations through relatively portable access (input--output, perturbation, gradients, activation statistics, etc.) into three classes (boundaries may overlap).

First, explanation by simplification approximates a complex model with simpler interpretable representations or surrogate models and explains through that approximation.
LIME and related methods are representative examples~\cite{arrieta2020}.

Second, feature attribution analysis assigns how much each feature or intermediate representation contributed to a prediction or a downstream representation.
Beyond input-dimension importance alone, the same framework can treat which upstream representations formed a given representation and which downstream representations it influenced.
Representative methods include SHAP-like approaches~\cite{arrieta2020}, Layer-wise Relevance Propagation (LRP)~\cite{bach2015,montavon2019}, and Integrated Gradients (IG)~\cite{sundararajan2017}.
For Transformers, AttnLRP~\cite{achtibat2024attnlrp}, which redistributes relevance along attention and matrix products, connects to this line.
IG assigns contributions by integrating gradients along a path from baseline to input; LRP backpropagates relevance from output to input and allocates it layer by layer.

Third, network analysis directly studies activation patterns, weight matrices, attention mechanisms, geometry of latent representations, etc., to reveal functions of particular units or subnetworks.
Circuit analysis targeting grammatical structure or multi-step reasoning also falls in this frame.
Attention Rollout~\cite{abnar2020}, which aggregates inter-layer attention matrices to analyze information flow, is a representative example of this view.

This work sits in feature attribution analysis above, and in particular uses set-to-set IG at within-layer module boundaries together with LRP-inspired path composition to analyze agreement between module-wise and layer-whole attribution (Sec.~\ref{sec:method}).

\subsection{Existing Transformer Interpretations and Limits}
\label{sec:related_transformer}

Attention Map visualization~\cite{vig2019,clark2019} is standard for Transformer interpretation.
However, attention weights need not match contributions to the output.
A low correlation between attention scores and sensitivity-based feature importance, as well as cases where randomizing attention scores barely changes accuracy, are reported in~\cite{jain2019}.
Query--Key similarity alone cannot fully explain outputs shaped by value vectors and the nonlinear MLP.

Attention Rollout~\cite{abnar2020} aggregates information flow by multiplying inter-layer attention matrices.
It can be read as factor attribution or as network analysis of cross-layer flow.
In either case it depends only on attention weights and ignores value magnitudes and MLP transforms.
Norm-based analysis using value norms as well as weights is proposed in~\cite{kobayashi2020}, showing that attention outputs depend on both.
Neither directly designs contributions that separate ATT and MLP within the same layer.

\subsection{Positioning of This Work}
\label{sec:related_positioning}

Most attribution methods surveyed in~~\cite{arrieta2020} (Saliency, DeepLIFT, SHAP, LIME) target scalar-valued objectives and are not naturally suited to baseline-controlled set-to-set comparison at ATT/MLP boundaries.
%Explanation can be applied at ATT/MLP boundaries within a layer, but methods surveyed by Arrieta et al.~\cite{arrieta2020}, such as Saliency, DeepLIFT, SHAP, and LIME, do not fit the design axis of this paper as naturally, because they rely on endpoint sensitivity, local linear approximation, or heavier computation rather than direct baseline comparison on ATT/MLP set-to-set maps.

The AttnLRP method~\cite{achtibat2024attnlrp} requires redesigning rules and references per operation and layer, imposing cost whenever the model or implementation changes.
Attention Map~\cite{vig2019,clark2019} and Rollout~\cite{abnar2020} rely on attention weights, ignoring values and MLP and known not to match contributions.
Circuit-Tracer~\cite{anthropic2025} analyzes circuits on interpretable SAE features rather than original representations.

By contrast, this paper applies set-to-set IG at module boundaries on the dynamic graph of Fig.~\ref{fig:intro-node-viz} and chains within-layer composition across those boundaries in an LRP-inspired way (Sec.~\ref{sec:method}).
Gradient-based attribution to intermediate layers exists (e.g., intermediate-layer DeepLIFT), but our novelty is twofold: IG is applied to set-to-set maps between token representation sets with baseline control at ATT/MLP boundaries, not only to a scalar classification objective; and separated ATT and MLP contributions are composed within a layer so that path-wise flow can be compared with layer-whole attribution.

\section{Transformer Graph View and Definitions}
\label{sec:transformer_graph}

This section defines per-layer computation in Transformers to clarify input--output variables and weight matrices in the attention mechanism and MLP (position-wise feed-forward network), as the basis for attribution with LIG (Layer-wise Integrated Gradients).

\subsection{Variables and Notation}
Let sequence length be $N$, model dimension $D$, and number of heads $H$.
Each head has dimension $D/H$.
We use BERT Base (Uncased), so $L{=}12$, $H{=}12$, $D{=}768$.
Let the input token set at layer $l$ be $z^{(l)} = \{z_i^{(l)}\}_{i=1}^N$, where $z_i^{(l)} \in \mathbb{R}^D$ is the representation of token $i$.
$l{=}0$ is the embedding-layer output; $l{=}1,\ldots,L{-}1$ correspond to outputs of each Transformer layer.
Multi-Head Attention (MHA) and attention-module notation in the literature are unified under ATT as in Sec.~\ref{sec:introduction}.
Below we write attention mechanism in definitions and explanations and $\mathrm{ATT}$ / $\mathrm{IG}^{\mathrm{ATT}}$ in maps, contributions, and paths (z2z, etc.).

Throughout Sec.~3 and Sec.~4, we omit the layer superscript $(l)$ and the head superscript $h$ when context fixes them, writing e.g. $z_i,u_i$ for $z_i^{(l)}, u_i^{(l)}$. We retain superscripts only when crossing layers or contrasting heads.

\subsection{Attention Mechanism}
%Fix layer $l$ and head $h$.
%Within the same layer and head we omit subscripts $(l)$ and $(l,h)$ and write $z_i,q_i,k_i,v_i,\alpha_{ij},u_i$.
%We distinguish $z^{(l)}$ and $z^{(l+1)}$ only when crossing layers.
%We also omit token index $i$ when clear from context, except when $i$ remains as an index over the full token set $\{z_i\}_i$ or in maps $\mathrm{ATT}(\{z_i\})=u_i$.
From input $z_i \in \mathbb{R}^D$, linear transforms yield Query $q_i$, Key $k_i$, and Value $v_i$ (each in $\mathbb{R}^{D/H}$):
\begin{align}
  q_i &= W_Q z_i + b_Q, \quad k_i = W_K z_i + b_K, \quad v_i = W_V z_i + b_V.
  \label{eq:att-qkv}
\end{align}
$W_Q,W_K,W_V$ are learned projection matrices from token representations to head dimension $D/H$; $b_Q,b_K,b_V$ are the corresponding biases (per layer $l$ and head $h$; written $W_Q^{(l,h)}$ etc.\ before omission).
Attention weights $\alpha_{ij}$ are defined by softmax over scaled dot products of query and key, and the head output $u_i$ is the value weighted average:
\begin{align}
  u_i &= \sum_{j=1}^{N} \alpha_{ij} v_j, \qquad
  \alpha_{ij} = \frac{\exp\bigl( (q_i \cdot k_j) / \sqrt{D/H} \bigr)}{\sum_{k=1}^{N} \exp\bigl( (q_i \cdot k_k) / \sqrt{D/H} \bigr)}.
  \label{eq:att-weight}
\end{align}
We call the matrix of $\alpha_{ij}$ on the right-hand side of Eq.~\eqref{eq:att-weight}, arranged as weights from query position $i$ to key position $j$, an Attention Map (visualizations of $\alpha_{ij}$ alone, without the value weighted average on the left-hand side).
$u_i \in \mathbb{R}^{D/H}$ is shorthand for attention output $u_i^{(l,h)}$ at token $i$ and head $h$; we write $\mathrm{ATT}(\{z_i\}_i) = u_i$ (strictly $\mathrm{ATT}^{(l,h)}$).
$\{z_i\}_i$ is the set of all token representations indexed by $i$; we sometimes omit $i$ and write $\mathrm{ATT}(\{z_i\}) = u_i$.

\subsection{MLP}
We describe the MLP for token $i$ at some layer $l$.
%Layer $l$ is fixed as above; we omit index $i$ when clear and write $o$ etc.\ (next-layer representation is $z_i^{(l+1)}$).
For token $i$, attention outputs $u_i^{(l,h)}$ are concatenated across heads, projected, and then passed through a residual connection and LayerNorm to obtain intermediate output $o_i$ (strictly $o_i^{(l)}$) as
\begin{align}
  o_i = \mathrm{LayerNorm}\left( z_i + W_o [u_i^{(l,1)}; \ldots; u_i^{(l,H)}] + b_o \right),
  \label{eq:mlp-o}
\end{align}
and the MLP body is two fully connected layers and activation $\phi$ (GELU in BERT).
The next-layer token representation is then given by
\begin{align}
  z_i^{(l+1)} = \mathrm{LayerNorm}\left( o_i + W_2 \phi(W_1 o_i + b_1) + b_2 \right),
  \label{eq:mlp-z}
\end{align}
where $W_o,W_1,W_2$ are learned weights of that layer; $b_o,b_1,b_2$ are biases.
We define $\mathrm{MLP}(\{u_i^{(l,h)}\}_h) = z_i^{(l+1)}$. %(strictly $\mathrm{MLP}^{(l)}$).
%$\{u_i^{(l,h)}\}_h$ is the set over heads $h$; with layer $l$ fixed we may omit $(l)$ and write $\{u_i^{(h)}\}_h$.

With layer $l$ fixed, within-layer computation is the composite $z^{(l)} \xrightarrow{\mathrm{ATT}} u \xrightarrow{\mathrm{MLP}} z^{(l+1)}$.
%Inputs depending on interpolation parameter $a$ are written $\{z_i(a)\}$, $\{u(a)\}$ with set indices omitted.
This work tracks contributions with $z$ and $u$ (strictly $z_i^{(l)}$ and $u_i^{(l,h)}$) as nodes (Fig.~\ref{fig:intro-node-viz}).

\subsection{Two Views of Within-Layer Contributions}
\label{sec:layer_two_views}
\begin{figure}[H]
  \centering
  \includegraphics[width=0.65\linewidth]{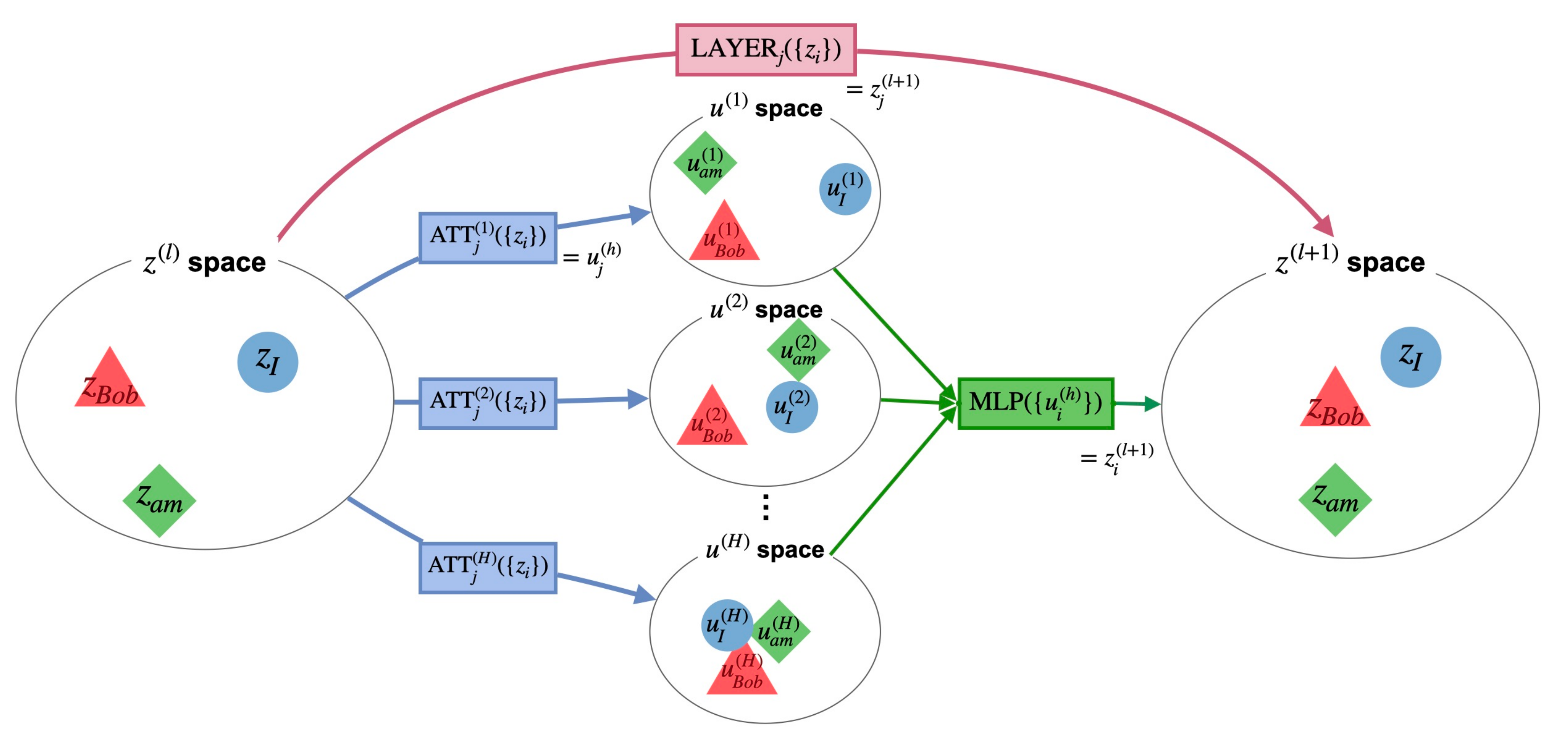}
  \caption{Two views of contributions within layer $l$.
  On the decomposed path below, IG measures $z^{(l)} \to u^{(l)}$ (z2u, ATT) and $u^{(l)} \to z^{(l+1)}$ (u2z, MLP), multiplying per-head ATT and MLP contributions and summing over heads as the path composition.
  On the layer-whole z2z path above, IG directly measures normalized layer-whole contribution $\widetilde{\mathrm{IG}}^{\mathrm{LAYER}}$ on $z^{(l)} \to z^{(l+1)}$ (z2z) and evaluates agreement with the composition.}
  \label{fig:z2z_z2u_u2z_views}
\end{figure}

Within the same layer $l$, we verify whether decomposition into ATT and MLP aligns with layer-whole contributions via the two paths in Fig.~\ref{fig:z2z_z2u_u2z_views} (bottom: the path measured separately for ATT and MLP; top: the layer-whole z2z path measured directly).
In other words, this paper considers decomposition methods such that the path composition $\widehat{\mathrm{IG}}^{\mathrm{ATT\circ MLP}}$, obtained by multiplying per-head ATT and MLP contributions and summing over heads, approaches $\widetilde{\mathrm{IG}}^{\mathrm{LAYER}}$.

\section{Analyzing Within-Layer Flow with LIG}
\label{sec:method}

This section formalizes LIG (Layer-wise Integrated Gradients) as used in this paper.
LIG uses set-to-set IG at module boundaries and LRP-inspired path composition within a layer.
The LIG view is not specific to Transformers.
Each internal representation (layer inputs/outputs, token embeddings, etc.) is a node, and each computational module is a local map from a set of input embeddings to a set of output embeddings.
Even when the map is nonlinear, IG along an interpolation path plus scalarization defines a scalar contribution per edge.
Treating these contributions as coefficients of a linear combination in the LRP view enables propagation analysis across module boundaries in one framework.
Below we formalize this with BERT's attention mechanism (ATT), MLP, and layer-whole maps as examples.

%Sec.~\ref{sec:def_ig} defines IG, Sec.~\ref{sec:def_lrp} relates LIG to LRP, and Sec.~\ref{sec:ig_setup} defines IG application to Transformers (scalarization and interpolation).
%Sec.~\ref{sec:three_views} formalizes within-layer composition and consistency; Sec.~\ref{sec:baseline} defines baselines; we end with numerical approximation.

\subsection{Definition of Integrated Gradients}
\label{sec:def_ig}

The typical IG setting attributes input $x\in\mathbb{R}^n$ to a scalar objective $f:\mathbb{R}^n\to\mathbb{R}$ (one logit component, loss, etc.).
We first state this standard form, then application to ATT and MLP in this paper.

Consider map $f:\mathbb{R}^n\to\mathbb{R}$ from input $x=(x_1,\ldots,x_n)$ to $y=f(x)$.
Throughout this paper, for computational cost and implementation simplicity, following \cite{sundararajan2017} we use a straight path from baseline $x(0)$ to actual input $x$, and define IG by
\begin{align}
  \mathrm{IG}_d(x)&=(x_d-x_d(0))\int_0^1 \frac{\partial f(x(a))}{\partial x_d}\,da,
    \label{eq:ig_straight}
\end{align}
where $x(a)=x(0)+a\,(x-x(0)),\ a\in[0,1]$.
By completeness, the sum of IG over input dimensions equals the output difference $f(x)-f(x(0))$.

Within a layer, ATT and MLP are vector-valued maps $F:\mathbb{R}^n\to\mathbb{R}^m$ from input $\mathbb{R}^n$ to output $\mathbb{R}^m$ ($n$ and $m$ may differ; strictly, maps between sets of token representations on a sequence).
To apply standard IG $f:\mathbb{R}^n\to\mathbb{R}$, we map $F$ to scalar objective $g:\mathbb{R}^n\to\mathbb{R}$.
Literature also uses per-output-component IG treating each $F_k$ as $f^{(k)}$; we do not use that for main results.
To obtain token-to-token contributions $\mathrm{IG}_{i,j}$, we apply Eq.~\eqref{eq:ig_straight} with $g(x)=\phi(F(x))$ using $\phi$ such as the $L_2$ norm of deviation from a reference point, in addition to interpolation and baseline design.
For attention, $\mathcal{A}_{j}$ in Eq.~\eqref{eq:ig-att}; for MLP, $\mathcal{M}^{(l)}$ in Eq.~\eqref{eq:ig-mlp}; for the layer, $\mathcal{L}_{j}$ in Eq.~\eqref{eq:ig-layer} correspond to this $g$.
The $L_2$ norm summarizes output-vector change in one scalar; completeness in Eq.~\eqref{eq:ig_straight} makes $g(x)-g(x(0))$ equal the sum of IG over input dimensions.

\begin{figure}[tb]
  \centering
  \includegraphics[width=0.50\linewidth]{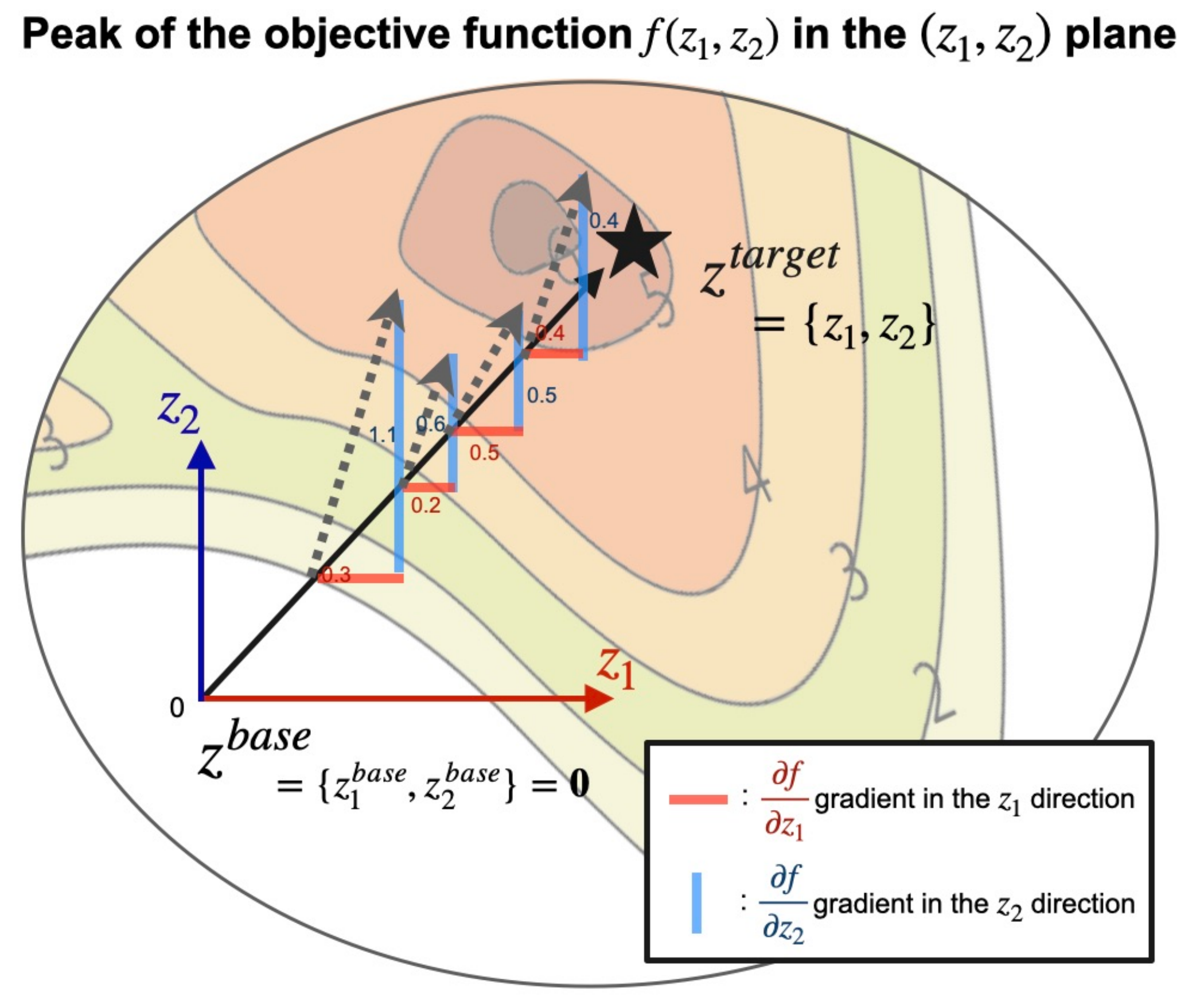}
  \caption{Conceptual diagram of Integrated Gradients (IG). Starting from baseline input $(z_1^{\mathrm{base}}, z_2^{\mathrm{base}})$ and moving along the straight path to actual input $(z_1, z_2)$, IG line-integrates the gradient at each point to quantify how much each input dimension $z_1, z_2$ contributes to the output difference $f(z_1, z_2)-f(z_1^{\mathrm{base}}, z_2^{\mathrm{base}})$.}
  \label{fig:ig_concept}
\end{figure}

\subsection{Relation to Layer-wise Relevance Propagation}
\label{sec:def_lrp}

Layer-wise Relevance Propagation (LRP)~\cite{bach2015,montavon2019} propagates relevance so that conservation,
\begin{align}
  \sum_i R_i^{(l)} = \sum_j R_j^{(l+1)},
\end{align}
is maintained across layers. In this paper we do not directly use rule-based propagation such as LRP-$\varepsilon$. Instead, we use IG on nonlinear module boundaries, where IG completeness gives the same kind of boundary-level completeness relation between scalarized output change and summed input contributions. In this sense, IG completeness plays at module boundaries the role that the conservation relation plays across layers in LRP, while within-layer composition follows the LRP spirit of layerwise propagation.%In that sense, IG serves here as a natural generalization of LRP-style completeness at module boundaries, while within-layer composition is treated in the LRP spirit as path composition.

\subsection{Applying IG to Transformers}
\label{sec:ig_setup}

Below we define scalarization of $\mathcal{A}_{j}$ (ATT), $\mathcal{M}^{(l)}$ (MLP), and $\mathcal{L}_j$ (layer-whole), and IG based on Eq.~\eqref{eq:ig_straight}.

To measure contribution to output token $j$'s representation $u_j$, for interpolation parameter $a \in [0,1]$ and interpolated input $z_i(a) = z_i^{\mathrm{base}} + a(z_i - z_i^{\mathrm{base}})$, we apply Eqs.~\eqref{eq:att-qkv}--\eqref{eq:att-weight} of Sec.~\ref{sec:transformer_graph} as written, with $z_i$ replaced by $z_i(a)$; the attention output is
\begin{align}
  \mathrm{ATT}_{j}(a) &= \mathrm{ATT}_{j}(\{z_i(a)\}_{i}) = u_{j}(a).
\end{align}
Using the $L_2$-norm scalarization $\mathcal{A}_{j}(a)$ for output token $j$, we apply Eq.~\eqref{eq:ig_straight} to $z_i \in \mathbb{R}^D$ (Sec.~\ref{sec:transformer_graph}).
Token $i$'s contribution is a scalar obtained by summing over input dimensions $d$:
\begin{align}
  \mathrm{IG}_{i,j,h}^{\mathrm{ATT}} &= \sum_{d=1}^{D}\left(z_{i,d}-z_{i,d}(0)\right)\int_0^1 \frac{\partial \mathcal{A}_{j}(a)}{\partial z_{i,d}}\,da,
  \qquad \mathcal{A}_{j}(a)=\|u_{j}(a)-u_{j}(0)\|_{2}.
  \label{eq:ig-att}
\end{align}
To make decomposition coefficients comparable across modules, we also use normalized contributions
\begin{align}
  \widetilde{\mathrm{IG}}_{i,j,h}^{\mathrm{ATT}} &= \frac{\mathrm{IG}_{i,j,h}^{\mathrm{ATT}}}{\mathcal{A}_{j}(1)-\mathcal{A}_{j}(0)},
  \qquad \sum_i \widetilde{\mathrm{IG}}_{i,j,h}^{\mathrm{ATT}} = 1.
  \label{eq:ig-att-norm}
\end{align}
Because each $\mathrm{IG}_{i,j,h}^{\mathrm{ATT}}$ is scalar, $\sum_i \widetilde{\mathrm{IG}}_{i,j,h}^{\mathrm{ATT}} = 1$ follows from IG completeness over tokens $i$ and dimensions $d$.

\label{sec:mlp_scalar_interp}
To measure contribution to next-layer input $z^{(l+1)}$, replace $u_i^{(l,h)}$ by interpolated input $u_i^{(l,h)}(a) = u_i^{(l,h),\mathrm{base}} + a(u_i^{(l,h)} - u_i^{(l,h),\mathrm{base}})$ in Eqs.~\eqref{eq:mlp-o} and \eqref{eq:mlp-z} of Sec.~\ref{sec:transformer_graph}.
Under this interpolation, MLP output is written as
\begin{align}
  \mathrm{MLP}^{(l)}(a) &= \mathrm{MLP}^{(l)}(\{u_{j}^{(l,h)}(a)\}_h) = z_{j}^{(l+1)}(a).
\end{align}
\begin{align}
  \mathrm{IG}_{h,j}^{\mathrm{MLP}} &= \sum_{d=1}^{D/H}\left(u_{j,d}^{(l,h)}-u_{j,d}^{(l,h)}(0)\right)\int_0^1 \frac{\partial \mathcal{M}^{(l)}(a)}{\partial u_{j,d}^{(l,h)}}\,da,
  \label{eq:ig-mlp}
\end{align}
where $\mathcal{M}^{(l)}(a)=\|z_{j}^{(l+1)}(a)-z_{j}^{(l+1)}(0)\|_{2}$.
Likewise, for headwise aggregation we use normalized MLP contributions $\widetilde{\mathrm{IG}}_{h,j}^{\mathrm{MLP}}$ so that $\sum_h \widetilde{\mathrm{IG}}_{h,j}^{\mathrm{MLP}} = 1$.
The definition is the same normalization form as Eq.~\eqref{eq:ig-att-norm}, replacing ATT quantities with the MLP scalarization $\mathcal{M}^{(l)}$.
Likewise, the layer-whole map $z^{(l)} \to z^{(l+1)}$ is written as
\begin{align}
  \mathrm{LAYER}_{j}^{(l)}(a) &= \mathrm{LAYER}_{j}^{(l)}(\{z_{i}(a)\}_{i}) = z_{j}^{(l+1)}(a).
\end{align}
\begin{align}
  \mathrm{IG}_{i,j}^{\mathrm{LAYER}} &= \sum_{d=1}^{D}\left(z_{i,d}-z_{i,d}(0)\right)\int_0^1 \frac{\partial \mathcal{L}_{j}(a)}{\partial z_{i,d}}\,da,
  \label{eq:ig-layer}
\end{align}
where $\mathcal{L}_{j}(a)=\|z_{j}^{(l+1)}(a)-z_{j}^{(l+1)}(0)\|_{2}$.
Likewise, we normalize layer-whole contributions by
\begin{align}
  \widetilde{\mathrm{IG}}_{i,j}^{\mathrm{LAYER}} &= \frac{\mathrm{IG}_{i,j}^{\mathrm{LAYER}}}{\mathcal{L}_{j}(1)-\mathcal{L}_{j}(0)},
  \qquad \sum_i \widetilde{\mathrm{IG}}_{i,j}^{\mathrm{LAYER}} = 1.
  \label{eq:ig-layer-norm}
\end{align}

\subsection{Within-Layer Composition and Consistency}
\label{sec:three_views}

For each layer $l$, head $h$, and output token $j$, we compute $\widetilde{\mathrm{IG}}_{i,j,h}^{\mathrm{ATT}}$ via Eq.~\eqref{eq:ig-att-norm}, normalized $\widetilde{\mathrm{IG}}_{h,j}^{\mathrm{MLP}}$ as defined just above, and $\widetilde{\mathrm{IG}}_{i,j}^{\mathrm{LAYER}}$ via Eq.~\eqref{eq:ig-layer-norm}.

Collecting contributions along headwise paths $z^{(l)} \to u^{(l,h)} \to z^{(l+1)}$, we write
\begin{align}
  \widehat{\mathrm{IG}}_{i,j}^{\mathrm{ATT\circ MLP}} &= \sum_{h=1}^{H}\left(\widetilde{\mathrm{IG}}_{i,j,h}^{\mathrm{ATT}}\cdot\widetilde{\mathrm{IG}}_{h,j}^{\mathrm{MLP}}\right),
  \qquad \widehat{\mathrm{IG}}_{i,j}^{\mathrm{ATT\circ MLP}} \approx \widetilde{\mathrm{IG}}_{i,j}^{\mathrm{LAYER}}.
  \label{eq:layer-decomp-hat}
\end{align}
Because $\sum_i \widetilde{\mathrm{IG}}^{\mathrm{ATT}}_{i,j,h} = 1$ for each $h$, the composition satisfies $\sum_i \widehat{\mathrm{IG}}^{\mathrm{ATT}\circ\mathrm{MLP}}_{i,j} = \sum_h \widetilde{\mathrm{IG}}^{\mathrm{MLP}}_{h,j} = 1$, so the chain of module boundaries preserves the conservation that LRP maintains across layers.
The path is $i \to \{h,j\} \to \{j\}$, and Eq.~\eqref{eq:layer-decomp-hat} contracts $h$; ``$\approx$'' indicates decomposition consistency between module-wise composition and layer-whole attribution.
($\widehat{\mathrm{IG}}^{\mathrm{ATT\circ MLP}}$ denotes the set obtained by summing headwise products over heads.)
Below we define baselines for each module (comparisons in Sec.~\ref{sec:experiments}).

\subsection{Baseline Definitions}
\label{sec:baseline}

Baseline design in this paper controls agreement with layer-whole attribution. Because MLP input is ATT output, our consistency-oriented MLP reference inherits the output at $a=0$ from the ATT-side path that uses the output token itself as the reference, rather than choosing an independent baseline. We define this configuration below as $\mathrm{ATTITBa}=0$.

With the zero vector (zero),
\begin{align}
  z_i(0)=\mathbf{0},\quad
  \mathrm{IG}_{i,j,h}^{\mathrm{ATT,zero}}=\sum_{d=1}^{D} z_{i,d}\int_0^1 \frac{\partial \mathcal{A}_{j}(a)}{\partial z_{i,d}}\,da.
\end{align}

With Input Token Base (ITB; self-input-token reference),
\begin{align}
  z_i(0)=z_{j},\quad \mathrm{IG}_{i,j,h}^{\mathrm{ATT,ITB}}=\sum_{d=1}^{D}(z_{i,d}-z_{j,d})\int_0^1 \frac{\partial \mathcal{A}_{j}(a)}{\partial z_{i,d}}\,da.
\end{align}
For $i=j$, $z_i(0)=z_j$ gives $(z_i-z_i(0))=0$, so Eq.~\eqref{eq:ig-att} implies that the self term $\mathrm{IG}_{j,j,h}^{\mathrm{ATT,ITB}}$ is theoretically zero~\cite{sturmfels2020}.
We estimate the self term in two ways.
Zero Base ratio (zeroRatio):
\begin{align}
\mathrm{IG}_{j,j,h}^{\mathrm{ATT,ITB,zeroRatio}}
&=
\mathrm{IG}_{j,j,h}^{\mathrm{ATT,zero}}
\cdot
\frac{
  \sum_{k\neq j} \mathrm{IG}_{k,j,h}^{\mathrm{ATT,ITB}}
}{
  \sum_{k\neq j} \mathrm{IG}_{k,j,h}^{\mathrm{ATT,zero}}
}
\end{align}
AttentionMap ratio (mapRatio):
\begin{align}
\mathrm{IG}_{j,j,h}^{\mathrm{ATT,ITB,mapRatio}}
&\approx
\left(
  \sum_{k\neq j} \mathrm{IG}_{k,j,h}^{\mathrm{ATT,ITB}}
\right)
\cdot
\frac{\alpha_{j,j}}{1-\alpha_{j,j}}
\end{align}
For $r\in\{\mathrm{zeroRatio},\mathrm{mapRatio}\}$, rescaling that preserves completeness:
\begin{align}
\widehat{\mathrm{IG}}_{i,j,h}^{\mathrm{ATT,ITB},r}
&=
\frac{
  \left\|u_{j}(1)-u_{j}(0)\right\|_2
}{
  \sum_{k\neq j}\mathrm{IG}_{k,j,h}^{\mathrm{ATT,ITB}}
  +
  \mathrm{IG}_{j,j,h}^{\mathrm{ATT,ITB},r}
}
\cdot
\begin{cases}
\mathrm{IG}_{j,j,h}^{\mathrm{ATT,ITB},r} & (i=j),\\
\mathrm{IG}_{i,j,h}^{\mathrm{ATT,ITB}} & (i\neq j)
\end{cases}
\end{align}
Ratio completion does not change ATT interpolation itself or $u_j(0)$.
For MLP (u2z), residual $z_i$ in Eq.~\eqref{eq:mlp-o} is not interpolated.
With the zero vector,
\begin{align}
  u(0)=\mathbf{0},\quad
  \mathrm{IG}_{h,j}^{\mathrm{MLP,zero}}=\sum_{d=1}^{D/H} u_{j,d}^{(l,h)}\int_0^1 \frac{\partial \mathcal{M}^{(l)}(a)}{\partial u_{j,d}^{(l,h)}}\,da.
\end{align}

\label{sec:attitba_zero_mlp}
For MLP (u2z), $\mathrm{ATTITBa}=0$ uses the ITB path on the ATT side and takes its endpoint output at $a=0$ as the MLP baseline.
A pure MLP-ITB that independently sets a self-token reference on MLP input also yields a trivially zero self term for the same reason and is unsuitable for contribution decomposition.

Under $\mathrm{ATTITBa}=0$,
\begin{align}
  z_k^{(l)}(a; j)=z_j^{(l)}+a\left(z_k^{(l)}-z_j^{(l)}\right),\quad
  u_j^{(l,h)}(0)=\mathrm{ATT}_j^{(l,h)}\!\left(\{z_k^{(l)}(0; j)\}_k\right),\\
  \mathrm{IG}_{h,j}^{\mathrm{MLP},\mathrm{ATTITBa}=0}=\sum_{d=1}^{D/H}\left(u_{j,d}^{(l,h)}-u_{j,d}^{(l,h)}(0)\right)\int_0^1 \frac{\partial \mathcal{M}^{(l)}(a)}{\partial u_{j,d}^{(l,h)}}\,da.
\end{align}
This keeps ATT and MLP on the same reference sequence and improves consistency with within-layer composition (Eq.~\eqref{eq:layer-decomp-hat}).

For layer-whole attribution (z2z), we use Zero / ITB / LAYER-ITB-zeroRatio.
LAYER-ITB-zeroRatio applies IG to the whole-layer map $z^{(l)}\!\to\! z^{(l+1)}$.
It uses ITB as the baseline and completes the self term by the ATT zeroRatio scheme.

\subsection{Numerical Approximation and Evaluation Metrics}

Implementation approximates the integral with $m$ segments:
\begin{align}
  \mathrm{IG}_{i,j}\approx (x_i-x_i(0))\cdot \frac{1}{m}\sum_{l=1}^{m} \left.\frac{\partial f(x(a))}{\partial x_i}\right|_{a=\frac{l}{m}}.
\end{align}
We typically use $m=32$; decomposition-consistency procedure and baseline combinations appear in Sec.~\ref{sec:experiments}.

\section{Experiments}
\label{sec:experiments}

\subsection{Setup}
We use pretrained BERT-base-uncased (12 layers, 12 heads).
Evaluation data are PTB dev sentences from Treebank-3~\cite{marcus1999treebank} (Stanford Dependencies format; indices $0$--$1699$, 1700 sentences) used for within-layer $L_2$ consistency analysis.

Flow-consistency analysis.
Following Eq.~\eqref{eq:layer-decomp-hat}, the first analysis asks how close the path composition of ATT-IG and MLP-IG is to reference layer-whole IG (z2z).
For each sentence, layer $l$, and output token $j$, let $\widetilde{\mathbf{c}}^{(l),\mathrm{LAYER}}_{:,j}$ and $\hat{\mathbf{c}}^{(l)}_{:,j}$ denote the vectors of normalized contributions over input tokens $i$ on the layer-whole z2z side and on the path-composed side, respectively.
The layer-whole side uses $\widetilde{\mathrm{IG}}^{\mathrm{LAYER}}_{i,j}$ (Eq.~\eqref{eq:ig-layer-norm}); the composed side uses $\sum_h (\widetilde{\mathrm{IG}}^{\mathrm{ATT}}_{i,j,h}\cdot\widetilde{\mathrm{IG}}^{\mathrm{MLP}}_{h,j})$ (Eq.~\eqref{eq:layer-decomp-hat}), and
\begin{equation}
  d^{(l)}_j=\left\|\widetilde{\mathbf{c}}^{(l),\mathrm{LAYER}}_{:,j}-\hat{\mathbf{c}}^{(l)}_{:,j}\right\|_2 .
  \label{eq:layer-decomp-l2}
\end{equation}
Table~\ref{tab:decomp_top} reports mean $d^{(l)}_j$ over all evaluable $(\text{sentence},l,j)$ pairs on PTB dev.
Reference z2z baselines are three variants: Zero / ITB / LAYER-ITB-zeroRatio; the path-composed side compares multiple ATT/MLP baseline combinations.
Table~\ref{tab:decomp_top} reports results in three disjoint blocks; $L_2$ ranks and absolute values are meaningful only within the same layer-whole z2z reference, not across blocks.

\subsection{Consistency of Within-Layer Composition and Layer-Whole Attribution}
\label{sec:exp_decomp}
Smaller $d^{(l)}_j$ means the within-layer split into ATT and MLP more closely reproduces layer-whole contributions under this analysis criterion.

% Decomposition consistency: Top-3 smallest L2 per layer_ig_baseline_group
% Source: summary_layer_vs_composed_affine.csv (PTB dev 1700 sentences; normalized IG, Eq. layer-decomp-hat)
\begin{table}[tb]
\caption{Flow-consistency diagnostic for within-layer composition and layer-whole attribution.
L2 averages $d^{(l)}_j$ (Eq.~\eqref{eq:layer-decomp-l2}) between normalized layer-whole contributions $\widetilde{\mathrm{IG}}^{\mathrm{LAYER}}_{:,j}$ (Eq.~\eqref{eq:ig-layer-norm}) and path-composed $\hat{\mathbf{c}}_{:,j}$ (Eq.~\eqref{eq:layer-decomp-hat}) over all evaluable pairs.
Left column: baseline of layer-whole z2z used as the reference; middle two columns: baselines on the side formed by path-composing ATT-IG and MLP-IG.
Because there are many candidates, this table compares only the top three configurations with smallest L2 for each reference z2z baseline (Zero / ITB / LAYER-ITB-zeroRatio).
Ranks are compared only within the same reference.
z2z, z2u, and u2z abbreviate IG paths to the layer whole, Attention, and MLP respectively.}
\label{tab:decomp_top}
\centering
\scriptsize
\begin{tabular}{@{}lllrr@{}}
\toprule
\multicolumn{1}{c}{Layer-whole z2z (reference)} & \multicolumn{2}{c}{ATT-IG $\times$ MLP-IG (path composition)} & L2 & Rank \\
\cmidrule(lr){2-3}
  & ATT baseline & MLP baseline &  &  \\
\midrule
\multicolumn{5}{@{}l}{\textit{Reference z2z: Zero}} \\
Zero & ITB-zeroRatio & Zero & 0.769 & 1 \\
Zero & ITB-zeroRatio & ATTITBa$=$0 & 0.770 & 2 \\
Zero & Zero & Zero & 0.777 & 3 \\
\midrule
\multicolumn{5}{@{}l}{\textit{Reference z2z: ITB (input token)}} \\
ITB & ITB & Zero & 0.214 & 1 \\
ITB & ITB & ATTITBa$=$0 & 0.214 & 2 \\
ITB & ITB-zeroRatio & Zero & 0.239 & 3 \\
\midrule
\multicolumn{5}{@{}l}{\textit{Reference z2z: LAYER-ITB-zeroRatio}} \\
LAYER-ITB-zeroRatio & ITB-zeroRatio & Zero & 0.784 & 1 \\
LAYER-ITB-zeroRatio & ITB-zeroRatio & ATTITBa$=$0 & 0.784 & 2 \\
LAYER-ITB-zeroRatio & Zero & Zero & 0.791 & 3 \\
\bottomrule
\end{tabular}
\end{table}

Table~\ref{tab:decomp_top} must not be read as a single global ranking by the $L_2$ column: absolute values differ markedly across reference blocks because the layer-whole baseline fixes both the reference vector and the $L_2$ scale.
Because ATT/MLP baseline combinations are numerous, including different self-term completions and MLP references, Table~\ref{tab:decomp_top} compares only the three configurations with smallest $L_2$ for each reference z2z baseline.
When the reference z2z baseline is ITB, $\mathrm{ATT}(\mathrm{ITB}) \times \mathrm{MLP}(\mathrm{Zero})$ is minimal at $L_2 = 0.214$. $\mathrm{ATT}(\mathrm{ITB}) \times \mathrm{MLP}(\mathrm{ATTITBa}{=}0)$ is nearly tied at $0.214$; both rank among the best in the ITB reference group.
As noted above, MLP input is ATT output. $\mathrm{ATTITBa}=0$ passes $u_j^{(l,h)}(0)$ at endpoint $a=0$ of the ATT ITB path as the MLP reference, inheriting the baseline sequence from ATT to MLP along within-layer flow. Using a zero-vector MLP baseline (Zero) is nearly as good (Table~\ref{tab:decomp_top}).
Under the normalized flow-consistency diagnostic, $\mathrm{MLP}(\mathrm{Zero})$ and $\mathrm{ATTITBa}=0$ best preserve agreement between module-wise decomposition and layer-whole attribution within the ITB reference group.
When the reference z2z baseline is Zero, $\mathrm{ATT}(\mathrm{ITB\text{-}zeroRatio}) \times \mathrm{MLP}(\mathrm{Zero})$ is minimal at $L_2 = 0.769$; $\mathrm{ATT}(\mathrm{ITB\text{-}zeroRatio}) \times \mathrm{MLP}(\mathrm{ATTITBa}=0)$ is nearly as good at $0.770$.
Across all three reference groups, changing only the MLP baseline between Zero and $\mathrm{ATTITBa}=0$ alters $L_2$ by at most $\approx 0.001$, whereas the ATT baseline determines the top-ranked configurations in Table~\ref{tab:decomp_top} within each group.

\subsection{Visualization with LIG: layer-wise token-to-token contributions}
\label{sec:exp_token_contrib}

\begin{figure}[tb]
  \centering
  \rotatebox{90}{%
    \includegraphics[height=0.60\linewidth]{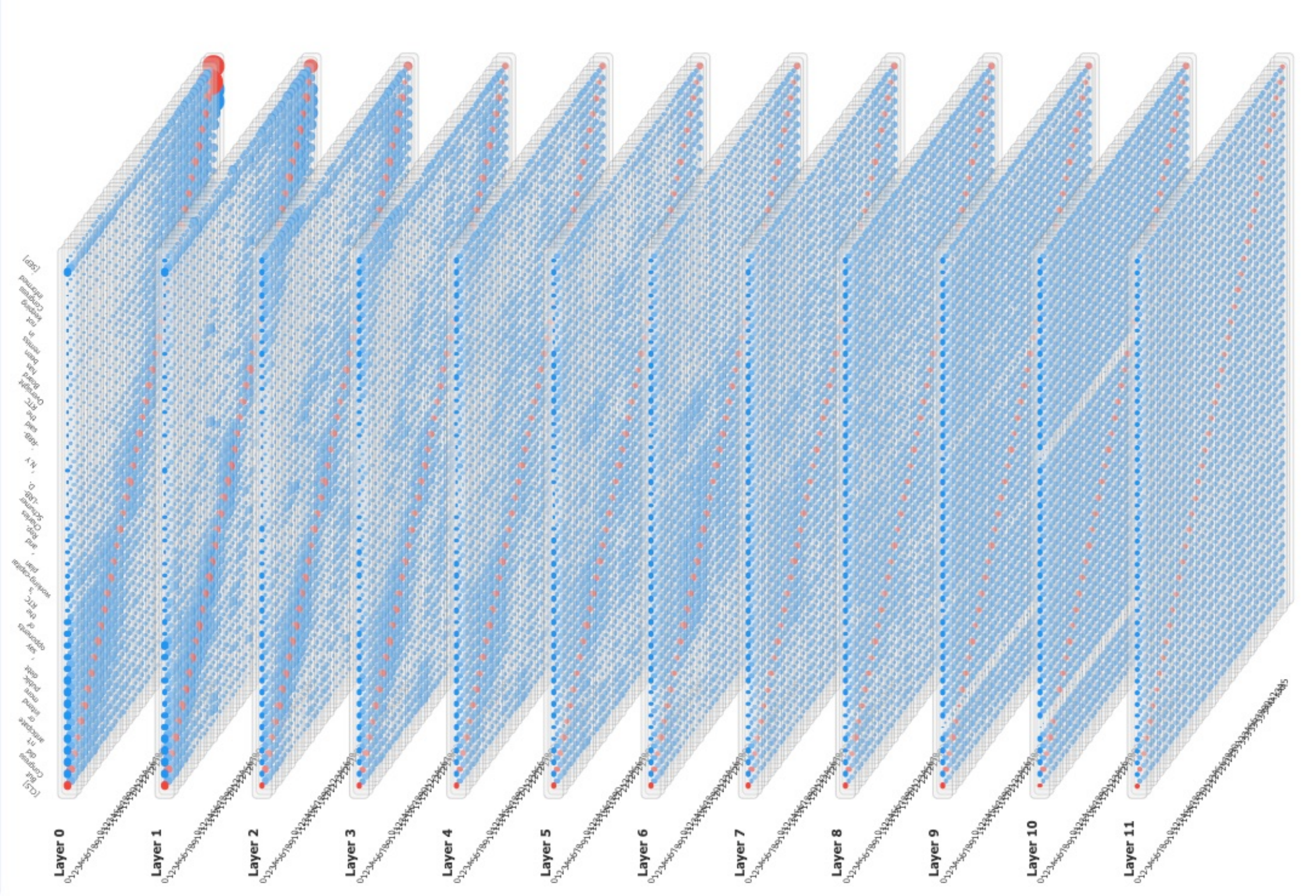}%
  }
  \caption{Layerwise token-to-token contribution visualization (z2z IG, Zero baseline).
  Layers progress left (Layer~0, embedding output) to right (Layer~11).
  Input sentence tokenized on whitespace:
  ``The firm's drop in net reflected weaker revenue in transactions for its own account -- a decline of 19\% to \$314.6 million on reduced revenue from trading fixed-income securities .''}
  \label{fig:z2z_token_contrib_example}
\end{figure}

At layer $l$, we apply Integrated Gradients (IG) to the layer-whole map $z^{(l)}\!\to\!z^{(l+1)}$ including Multi-Head Attention (MHA) and MLP, obtaining a matrix of contributions from each prior-layer token representation to each next-layer token representation (layer-whole z2z; Sec.~\ref{sec:method}).
Figure~\ref{fig:z2z_token_contrib_example} shows a qualitative example with z2z and zero-vector baseline: contribution matrices from Layer~0 through Layer~11 side by side.
In each panel the vertical axis is the prior layer (token symbols), the horizontal axis is the next layer (token indices), and circle size is contribution magnitude.
Red is contribution from the target token itself; blue is from other tokens.

Near the input (e.g.\ Layer~0), contributions concentrate on the token and its neighborhood; local relations between adjacent tokens (phrases) are strong.
Around Layer~4, contributions to expressions such as \textit{The firm's drop in net} increasingly come from distant tokens such as \textit{a decline of 19\%} that are semantically linked in the sentence.
From Layer~9 to Layer~11 the distribution becomes more uniform as information integration spans the whole sentence.
Thus, in this qualitative example, we read a layerwise division of roles: local relation organization early, organization of semantically or syntactically important relations at some distance in the middle, and sentence-wide integration late.
This reading is consistent with the local-to-global hierarchy reported for interpretations based on attention weights alone, but we do not quantify it here.

\section{Discussion}
\label{sec:discussion}

LIG provides a common criterion for comparing how closely ATT/MLP path decomposition matches layer-whole z2z attribution within a layer.
We analyze this through $L_2$ consistency.
ITB measures incremental contributions, but an independent ITB on MLP tends to cause inconsistency; $\mathrm{ATTITBa}=0$ reduces that mismatch by passing the ATT integration endpoint to MLP. Using Zero as the MLP reference is nearly as good.

Table~\ref{tab:decomp_top} suggests that $\mathrm{MLP\text{-}IG}$ is relatively insensitive to baseline choice between Zero and $\mathrm{ATTITBa}=0$, whereas $\mathrm{ATT\text{-}IG}$ baseline and self-term completion matter more.

Figure~\ref{fig:z2z_token_contrib_example} (Sec.~\ref{sec:exp_token_contrib}) qualitatively suggests hierarchical roles: local organization in shallow layers, long-range semantic and syntactic relations in the middle, and sentence-wide integration in deep layers.

Attention Maps can be strong per relation but ignore Value and MLP.
The proposed LIG therefore serves as an analysis tool: at module-boundary granularity it makes visible where module-wise flow aligns with layer-whole attribution.

Limitations.
(i)~Restricted to BERT-base and PTB dev sentences; because set-to-set IG at each module boundary and head is costly, we validate the framework on BERT-base, the lightest and most widely used standard encoder.
(ii)~Straight paths may pass through OOD regions~\cite{sanyal2021,hase2021}.
(iii)~Theoretical uniqueness of ITB completion (zeroRatio/mapRatio) remains future work.
(iv)~The local-to-global shift suggested in Fig.~\ref{fig:z2z_token_contrib_example} rests on a qualitative visualization only; quantifying depth-dependent processing roles remains future work.

\section{Conclusion}
\label{sec:conclusion}

This paper viewed a Transformer as a dynamic graph with $z$ and $u$ as nodes and used LIG to analyze within-layer flow.
Set-to-set IG measures contributions on nonlinear module boundaries, and IG completeness provides at those boundaries the same kind of completeness relation that motivates LRP-style conservation. Attention and MLP contributions are measured separately so that their path composition can be compared with layer-whole attribution.
The flow-consistency analysis showed that baseline design matters, in particular for $\mathrm{ATT\text{-}IG}$ and the layer-whole z2z reference. Inheriting the ATT integration endpoint as the MLP reference ($\mathrm{ATTITBa}=0$) ranks among the best normalized $L_2$ configurations; $\mathrm{MLP}(\mathrm{Zero})$ performs nearly as well once IG is normalized to a completeness-preserving allocation distribution.
Layer-wise visualization further illustrated how separated ATT/MLP flow composes and how processing roles evolve across depth.
In this sense, we present LIG as an XAI tool that applies set-to-set IG at module boundaries, diagnoses within-layer consistency ($L_2$) by comparing path composition with layer-whole attribution, and supports macroscopic reading of within-layer information flow.

Future work includes application to large LLMs, baseline guidance on other tasks, and study of path shape.

\bibliographystyle{splncs04}
\bibliography{references}

\begin{thebibliography}{10}
\providecommand{\url}[1]{\texttt{#1}}
\providecommand{\urlprefix}{URL }
\providecommand{\doi}[1]{https://doi.org/#1}

\bibitem{vaswani2017}
Vaswani, A., Shazeer, N., Parmar, N., Uszkoreit, J., Jones, L., Gomez, A.N.,
  Kaiser, {\L}., Polosukhin, I.: Attention is all you need. In: Advances in
  Neural Information Processing Systems (NeurIPS). pp. 5998--6008 (2017)

\bibitem{devlin2019}
Devlin, J., Chang, M.W., Lee, K., Toutanova, K.: {BERT}: Pre-training of deep
  bidirectional transformers for language understanding. In: Proceedings of
  NAACL-HLT. pp. 4171--4186 (2019)

\bibitem{brown2020}
Brown, T., Mann, B., Ryder, N., Subbiah, M., Kaplan, J., Dhariwal, P.,
  Neelakantan, A., Shyam, P., Sastry, G., Askell, A., et~al.: Language models
  are few-shot learners. In: Advances in neural information processing systems.
  vol.~33, pp. 1877--1901 (2020)

\bibitem{arrieta2020}
Arrieta, A.B., D{\'i}az-Rodr{\'i}guez, N., Del~Ser, J., Bennetot, A., Tabik,
  S., Barbado, A., Garc{\'i}a, S., Gil-L{\'o}pez, S., Molina, D., Benjamins,
  R., Chatila, R., Herrera, F.: Explainable artificial intelligence (xai):
  Concepts, taxonomies, opportunities and challenges toward responsible ai.
  Information Fusion  \textbf{58},  82--115 (2020)

\bibitem{bach2015}
Bach, S., Binder, A., Montavon, G., Klauschen, F., M{\"u}ller, K.R., Samek, W.:
  On pixel-wise explanations for non-linear classifier decisions by layer-wise
  relevance propagation. PLOS ONE  \textbf{10}(7),  e0130140 (2015)

\bibitem{montavon2019}
Montavon, G., Lapuschkin, S., Binder, A., Samek, W., M{\"u}ller, K.R.:
  Layer-wise relevance propagation: An overview. Explainable {AI}:
  Interpreting, Explaining and Visualizing Deep Learning pp. 193--209 (2019)

\bibitem{sundararajan2017}
Sundararajan, M., Taly, A., Yan, Q.: Axiomatic attribution for deep networks.
  Proceedings of the 34th International Conference on Machine Learning pp.
  3319--3328 (2017)

\bibitem{achtibat2024attnlrp}
Achtibat, R., Hatefi, S.M.V., Dreyer, M., Samek, W., Lapuschkin, S.: {AttnLRP}:
  {Attention-Aware} {Layer-Wise} {Relevance} {Propagation} for {Transformers}.
  In: Proceedings of the International Conference on Machine Learning (ICML)
  (2024)

\bibitem{abnar2020}
Abnar, S., Zuidema, W.: Quantifying attention flow in transformers. In:
  Proceedings of ACL. pp. 4190--4197 (2020)

\bibitem{vig2019}
Vig, J.: A multiscale visualization of attention in the transformer model. In:
  Proceedings of the 57th Annual Meeting of the Association for Computational
  Linguistics: System Demonstrations. pp. 37--42 (2019)

\bibitem{clark2019}
Clark, K., Khandelwal, U., Levy, O., Manning, C.D.: What does {BERT} look at?
  an analysis of {BERT}'s attention. In: Proceedings of the 2019 ACL Workshop
  BlackboxNLP. pp. 276--286 (2019)

\bibitem{jain2019}
Jain, S., Wallace, B.: Attention is not explanation. In: Proceedings of
  NAACL-HLT. pp. 3543--3556 (2019)

\bibitem{kobayashi2020}
Kobayashi, G., Kuribayashi, T., Yokoi, S., Inui, K.: Attention is not only a
  weight: Analyzing transformers with vector norms. In: Proceedings of the 2020
  Conference on Empirical Methods in Natural Language Processing (EMNLP). pp.
  7057--7075 (2020)

\bibitem{anthropic2025}
Hanna, M., Piotrowski, M., Lindsey, J., Ameisen, E.: Circuit-tracer: A new
  library for finding feature circuits. In: Proceedings of the 8th BlackboxNLP
  Workshop: Analyzing and Interpreting Neural Networks for NLP. pp. 239--249.
  Association for Computational Linguistics, Suzhou, China (2025)

\bibitem{sturmfels2020}
Sturmfels, P., Lundberg, S., Lee, S.I.: Visualizing the impact of feature
  attribution baselines. Distill  (2020)

\bibitem{marcus1999treebank}
Marcus, M.P., Santorini, B., Marcinkiewicz, M.A., Taylor, A.: Treebank-3. Web
  Download (1999), lDC Catalog No. LDC99T42.
  \url{https://catalog.ldc.upenn.edu/LDC99T42}. DOI:
  \url{https://doi.org/10.35111/gq1x-j780}

\bibitem{sanyal2021}
Sanyal, S., Ren, X.: Discretized integrated gradients for explaining language
  models. In: Proceedings of the 2021 Conference on Empirical Methods in
  Natural Language Processing (EMNLP). pp. 10285--10299 (2021)

\bibitem{hase2021}
Hase, P., Xie, H., Bansal, M.: The out-of-distribution problem in
  explainability and search methods for feature importance explanations. In:
  Advances in Neural Information Processing Systems (NeurIPS). vol.~34, pp.
  3650--3666 (2021)

\end{thebibliography}

\end{document}